\pdfoutput=1

\documentclass[11pt]{article}

\usepackage[final]{acl}

\usepackage{times}
\usepackage{latexsym}
\usepackage{booktabs}    
\usepackage{makecell}
\usepackage{graphicx} 
\usepackage{multirow} 
\usepackage[T1]{fontenc}

\usepackage[utf8]{inputenc}

\usepackage{microtype}

\usepackage{inconsolata}

\usepackage{graphicx}
\usepackage{amsmath}
\usepackage{amsfonts}
\usepackage{natbib}

\title{Beyond Position: the emergence of wavelet-like properties in Transformers}

\author{Valeria Ruscio, Umberto Nanni, Fabrizio Silvestri \\
        Sapienza University of Rome \\
        \texttt{ruscio@diag.uniroma1.it}, \texttt{fsilvestri@diag.uniroma1.it}}

\begin{document}
\maketitle

\begin{abstract}
This paper studies how Transformer models with Rotary Position Embeddings (RoPE) develop emergent, wavelet-like properties that compensate for the positional encoding's theoretical limitations. Through an analysis spanning model scales, architectures, and training checkpoints, we show that attention heads evolve to implement multi-resolution processing analogous to wavelet transforms. We demonstrate that this scale-invariant behavior is unique to RoPE, emerges through distinct evolutionary phases during training, and statistically adheres to the fundamental uncertainty principle. Our findings suggest that the effectiveness of modern Transformers stems from their remarkable ability to spontaneously develop optimal, multi-resolution decompositions to address inherent architectural constraints.


\end{abstract}
\section{Introduction}

Position encoding mechanisms are fundamental to Transformer architectures, enabling these inherently permutation-invariant models to capture sequential information. While early approaches relied on fixed sinusoidal encodings \citep{vaswani2017attention}, Rotary Positional Embeddings (RoPE) \citep{su2024roformer} represents a significant advancement by integrating relative positional information through rotational transformations. Despite RoPE's widespread adoption and empirical success, theoretical analysis suggests inherent limitations in its ability to simultaneously achieve high positional precision and frequency resolution \citep{barbero2024round}, a trade-off analogous to the uncertainty principle in signal processing. This creates a paradox: why do models with these theoretical constraints perform so well in practice?

We argue that this is resolved because RoPE-equipped models learn to compensate by developing emergent, wavelet-like processing strategies. Our analysis shows that Transformer attention heads do not act as a monolithic block; instead, they spontaneously organize into a multi-resolution framework. Different heads specialize in processing information at distinct frequency bands, effectively decomposing the input signal in a manner strikingly similar to a discrete wavelet transform.

Our work makes the following contributions:
\begin{itemize}
\item We demonstrate that the emergence of robust, wavelet-like, scale-invariant properties is a distinctive feature of the RoPE architecture compared to other common position encoding schemes.
\item We provide an evolutionary analysis of these properties, revealing the distinct learning phases models undergo to form their spectral processing strategies.
\item We offer a detailed statistical characterization of attention head behavior, showing that models learn to respect the physical constraints of the uncertainty principle and develop diverse internal strategies.
\end{itemize}

These findings reveal transformers' remarkable adaptability in discovering and implementing optimal solutions to complex information processing challenges. Appendices \ref{limitations}, \ref{linguistics}, and \ref{metrics} provide further detail on RoPE's limitations, the relationship between language and wavelets, and our metric definitions.

\section{Related Works}

The Transformer architecture \cite{vaswani2017attention} revolutionized sequence modeling through self-attention mechanisms. While the original Transformer used simple sinusoidal positional encodings, recent work has explored more sophisticated approaches. ALiBi \citep{press2021train} introduced attention bias terms that scale with relative position, while T5 \citep{raffel2020exploring} employed learned relative position embeddings. RoPE \citep{su2024roformer} advanced this further by applying rotation matrices to embeddings, though it faces fundamental limitations rooted in the uncertainty principle between position and frequency domains.

Neural networks' behavior, particularly their nonlinear components, has been increasingly analyzed through signal processing principles. Research has shown that activation functions can generate higher-order harmonics and exhibit frequency mixing \citep{selesnick1998generalized, rahimi2008weighted}, while principles of constructive and destructive interference have proven valuable in analyzing network behavior \citep{oppenheim1999discrete, chi2020fast}. Information-theoretic analyses of neural networks \citep{shwartz2017opening} have provided insights into their representational capabilities and limitations. Studies have examined how information flows through layers \citep{goldfeld2018estimating} and how architectural choices affect information bottlenecks \citep{tishby2015deep}. This theoretical framework has proven particularly valuable in understanding the capacity limitations of various neural network components.

\section{Methodology}
We integrate frequency-domain analyses, wavelet-based multi-scale decomposition, and entropy-based uncertainty assessments to comprehensively characterize the emergent properties of these models. Our methodology is designed to isolate positional encoding behaviors, assess their stability across model scales and architectures, and validate their alignment with theoretical expectations related to the trade-off between positional resolution and spectral organization.

\subsection{Frequency Analysis}
To probe the spectral properties of attention distributions, we employed a frequency-domain analysis using the Discrete Fourier Transform (DFT), we used the Hann window and zero padding. For each attention head $h$ within each model, we represented the attention pattern over token positions as $a_h(t)$, where $t$ indexes tokens within a single sequence. We computed the power spectral density (PSD):
\begin{equation}
    P_h(\omega) = |\mathcal{F}{a_h t}|^2
\end{equation}
where $\mathcal{F}$ denotes the DFT and $\omega$ the angular frequency. The frequency domain was partitioned into low (0-0.25 $\omega_N$), mid (0.25-0.75 $\omega_N$), and high (0.75-$\omega_N$) bands, where $\omega_N$ is the Nyquist frequency corresponding to the maximum resolvable frequency for the given sequence length.

The Nyquist frequency $\omega_N$  is set to half the sampling rate (1/2 tokens) for three fundamental reasons: it represents the highest meaningful frequency in discrete token sequences, as attention patterns can only alternate between consecutive tokens, making faster oscillations indistinguishable due to aliasing. Second, it provides natural normalization across sequence lengths, while absolute frequency ranges differ, all sequences share the same relative frequency structure when normalized by 
$\omega_N$, enabling meaningful cross-length comparisons of attention head frequency sensitivity. Third, following Shannon's sampling theorem, $\omega_N$ represents the theoretical maximum rate for information transmission through a discrete channel, thus defining the finest granularity at which positional information can be encoded without loss, making it the natural choice for analyzing models' representational capacity distribution.

To quantify the relative emphasis a head places on different frequency bands, we computed:
\begin{equation}
    \beta_h(b) = \frac{\int_b P_h(\omega)d\omega}{\int^{\omega N}_0 P_h(\omega)d\omega}
\end{equation}
where $b$ is the frequency band under consideration. 
To measure how selectively each attention head responds to specific frequencies, we define the frequency selectivity $S(h)$ for head $h$ as:
\begin{equation}
    S(h) = \frac{\max_{\omega}\{P_h(\omega)\}}{\int_0^{\omega_N} P_h(\omega)d\omega - \max_{\omega}\{P_h(\omega)\}}
\end{equation}
where $P_h(\omega)$ is the power spectral density at frequency $\omega$, and $\omega N$ is the Nyquist frequency, and a higher value indicates more focused frequency tuning of the head.

These frequency-domain analyses allowed us to discern how attention heads distribute their representational capacity across multiple scales, testing the premise that models spontaneously develop organized frequency content despite RoPE's intrinsic limitations.

\subsection{Wavelet Analysis}
While frequency-domain analysis captures global spectral properties, it lacks explicit positional localization. To address this, we employed wavelet decompositions using the Daubechies-2 (db2) wavelet \footnote{The selection of Daubechies-2 (db2) for our primary analysis reflects its optimal balance between smoothness and localization properties. Its compact support of length 4 aligns well with typical attention spans in language processing, while its vanishing moment enables effective detection of local linguistic changes against broader contextual background. The db2 wavelet's mathematical normalization properties match the normalization constraints imposed by softmax attention mechanisms, while its orthogonality properties prevent interference between shifted attention patterns.}. Wavelets offer a time-frequency (or position-frequency) representation that enables simultaneous assessment of spatial localization and scale-dependent behaviors.

For each head $h$, we computed wavelet coefficients:
\begin{equation}
    W_h(s,\tau) = \int a_h(t)\psi_{s,\tau}(t) dt
\end{equation}
where $\psi_{s,\tau}(t)$ is the mother wavelet at scale $s$ and translation $\tau$. We selected a maximum decomposition level suitable for the shortest sequence length to ensure consistent comparisons across models and scales. Wavelet entropy was computed at each scale:
\begin{equation}
    H_w(s) = - \sum_{\tau} |W_h(s, \tau)|^2 \log{(|W_h(s, \tau)|^2)}
\end{equation}
providing a measure of how the model distributes attention energy and complexity across different scales and positional shifts.

\subsection{Uncertainty Analysis}
To evaluate the theoretical trade-off between positional precision and spectral organization, we computed entropy measures for both the positional and spectral domains. Positional entropy $H_p(h)$ was derived from attention distributions over token positions:
\begin{equation}
    H_p(h) = - \sum_{\tau} a_h(t) \log{a_h(t)}
\end{equation}
reflecting how evenly attention is spread across the sequence. Similarly, spectral entropy $H_s(h)$ was computed from the normalized power spectrum $\hat{P}_h(\omega)$:
\begin{equation}
    H_s(h) = - \sum_{\omega} \hat{P}_h(\omega)\log{\hat{P}_h (\omega)}
\end{equation}
where $\hat{P}_h(\omega) = \frac{P_h(\omega)}{\sum_{\omega} P_h(\omega)}$ is the normalized power spectrum.

To quantify the relationship between these entropy measures, we define the position-spectrum correlation $\rho(h)$ through their normalized covariance:
\begin{equation}
    \rho(h) = \frac{\text{Cov}(H_p(h), H_s(h))}{\sigma_{H_p} \sigma_{H_s}}
\end{equation}

This correlation is then aggregated across all attention heads in a layer to measure how well the model balances the uncertainty principle trade-off between positional and spectral information:
\begin{equation}
    \rho_{\text{layer}} = \text{mean}_{h\in\text{layer}}\{\rho(h)\}
\end{equation}

The layer-wise correlation metric is bounded by $[-1, 1]$, with values closer to -1 indicating strong trade-offs between positional and spectral precision, and values closer to 1 indicating successful integration of both domains.

By comparing $H_p(h)$ and $H_s(h)$ through these correlation metrics, we can ascertain whether the model's attention patterns obey an uncertainty principle-like trade-off, wherein improved positional localization may come at the cost of reduced spectral complexity, or vice versa.

\subsection{Scale Invariance Testing}
We hypothesized that the models’ compensatory strategies would exhibit scale invariance properties, i.e., the ability to maintain positional-awareness structures when the input sequence length changes. To test this, we generated scaled variants $x_{\alpha}$ of each input sequence $x$ by sampling $\lfloor \alpha n \rfloor$ tokens, with $\alpha \in \{0.5, 0.25\}$ and $n$ the original sequence length. After computing the wavelet coefficients $W_h(x)$ and $W_h(x_{\alpha})$, we measured the scale sensitivity:
\begin{equation}
    S_h(\alpha) = 1-\cos({W_h(x), W_h(x_{\alpha}))}
\end{equation}
where $\cos{(\cdot, \cdot)}$ denotes cosine similarity. A low $S_h(\alpha)$ indicates that wavelet coefficients remain stable under rescaling, suggesting robust scale-invariant positional representations.
 
\subsection{Frame Completeness}
To verify that the learned representations form a stable, frame-like basis capable of faithful reconstruction, we performed inverse wavelet transforms. The reconstruction error $\varepsilon$ was computed as:
\begin{equation}
    \varepsilon = \frac{||a_h - W^{-1}(W_h) ||_F}{||a_h||_F}
\end{equation}
where $W^{-1}(\cdot)$ denotes the inverse wavelet transform and $||\cdot||_F$ is the Frobenius norm. A small $\varepsilon$ indicates that the attention patterns are well-represented by their wavelet coefficients, reinforcing the notion that the model’s positional strategies form a coherent, frame-like structure.

\section{Implementation Details}
We selected five pre-trained Transformer-based language models that vary in size, architecture, and training regimen to ensure the generality of our findings. Specifically, we analyzed Gemma 2 2B, Pythia 2.8B and 12B, LLaMA-3-2 1B, Mistral 7B, and Qwen 2.5 5B. These models encompass a wide parameter range (1B–12B), capturing different representational capacities and training protocols.

All models were evaluated on a curated sample of 500 sequences drawn from wikipedia. The selected sequences varied in length to expose scale-dependent behavior and stress-test the models’ positional encoding strategies under diverse conditions.

All experiments were conducted using PyTorch on A100, L4, and T4 GPUs to ensure computational efficiency and scalability. Frequency and spectral computations employed standard FFT-based routines, while wavelet transforms were performed using the PyWavelets library with a decomposition level chosen based on the minimum sequence length. Before analysis, attention weights were normalized and numerically stabilized to mitigate floating-point underflow, with a threshold of $10^{-10}$ applied to division operations.

\section{Experiments and Analysis}
Our analysis shows that Transformer models with RoPE spontaneously develop a sophisticated, multi-resolution processing strategy, similar to wavelet decomposition, to overcome the theoretical limitations of their position embeddings. This emergent behavior is not an isolated artifact but a consistent pattern substantiated by three key lines of evidence: the hierarchical organization of attention, quantitative analysis of model scaling and stability, and statistical confirmation of a fundamental signal processing principle.

\begin{figure}[h!]
\centering
\includegraphics[width=\linewidth]{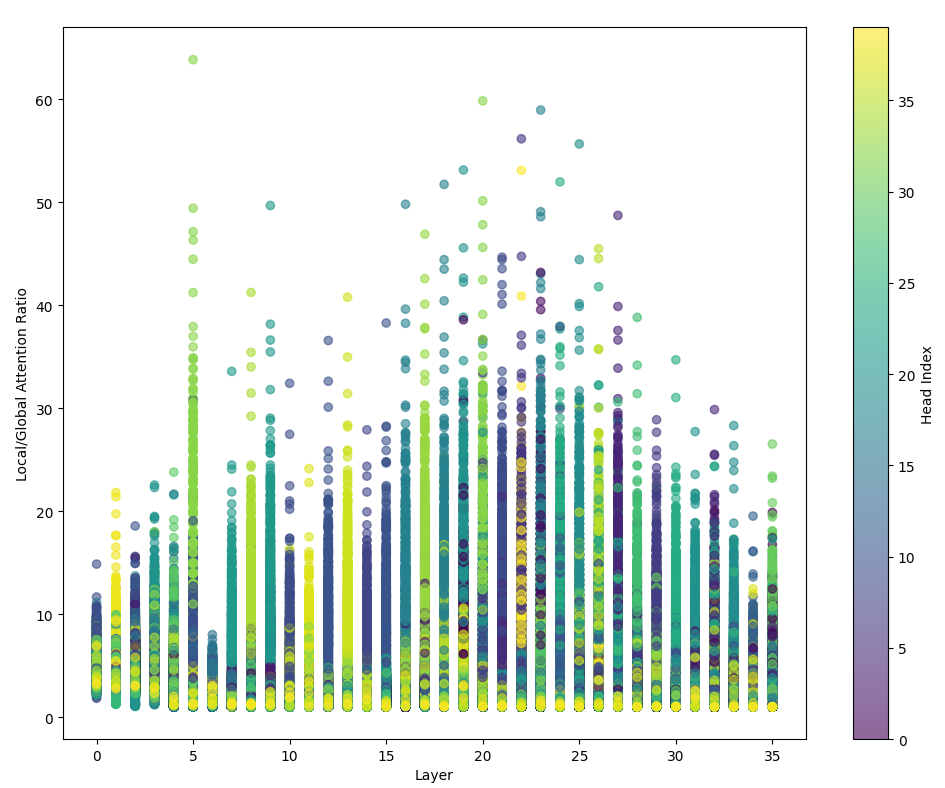}
\caption{Local vs Global attention distribution from Pythia 12B}
\label{fig:local-global}
\end{figure}

\paragraph{The Emergence of Multi-Scale Processing}
In figure \ref{fig:local-global} we can see that attention heads specialize into either local or global processors, evidenced by the pronounced vertical striping in visualizations of local-to-global attention ratios. This specialization deepens through the layers, with increasing variance that closely resembles the branching structure of a wavelet packet decomposition tree.

\begin{figure}[h!]
\centering
\includegraphics[width=\linewidth]{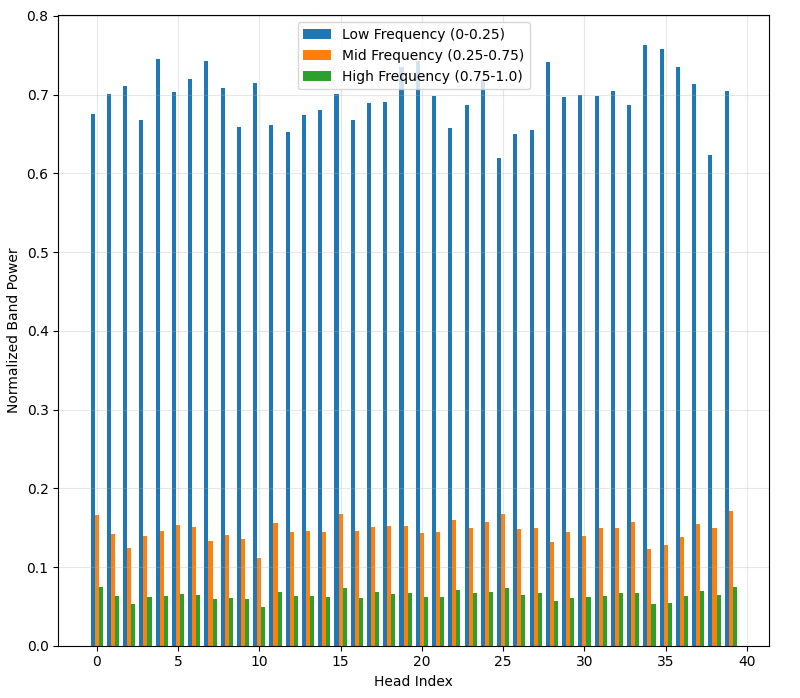}
\caption{Frequency band distribution across heads from Pythia 12B}
\label{fig:frec-band}
\end{figure}

\begin{figure}[h]
\centering
\includegraphics[width=\linewidth]{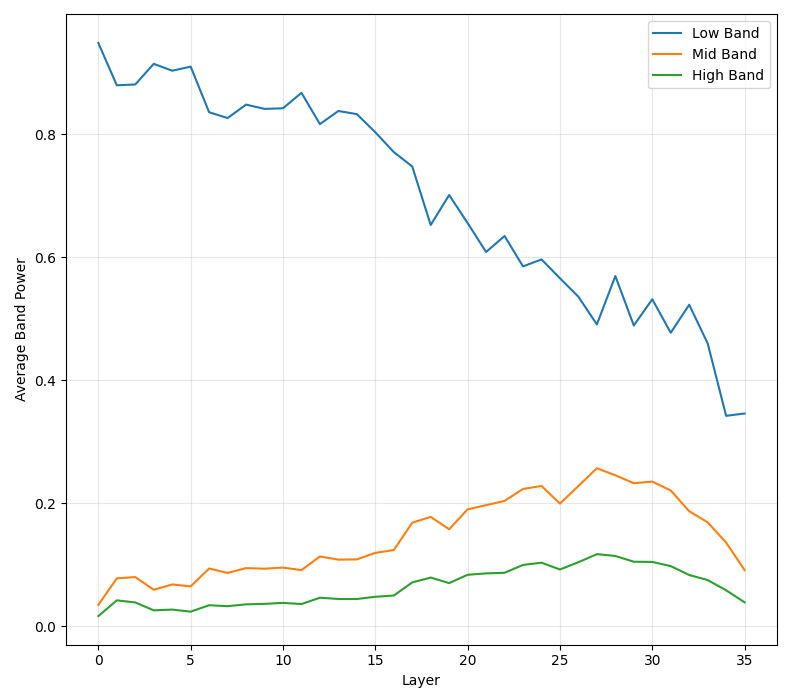}
\caption{Frequency response evolution across layers from Pythia 12B}
\label{fig:fre-respo}
\end{figure}

This multi-resolution strategy is also evident in the frequency domain. Analysis of attention patterns, as in figure \ref{fig:frec-band}, shows a consistently stratified frequency response. Low-frequency components (0–0.25 $\omega N$) form the contextual backbone, capturing 60–80\% of the spectral power. Mid-frequencies (0.25–0.75 $\omega N$) contribute a stable 15–25\%, while high-frequencies (0.75–$\omega N$) handle fine-grained details with a smaller 5–15\% share. In figure \ref{fig:fre-respo}, we can see that as information propagates through the model, we observe a smooth, dynamic evolution where the initial dominance of low frequencies tapers off, and mid- to high-frequency components gain influence, mirroring the adaptive refinement process of a wavelet analysis.

\subsection{Quantitative Analysis of Developmental Trajectories}

Quantitative metrics confirm these visual observations and reveal deeper patterns related to model scale and capacity, as we can see from table \ref{tab:main-metrics}. The model's inherent wavelet-like response is particularly evident in the scale sensitivity metric. As predicted by wavelet theory, sensitivity to sequence rescaling is not only low but also degrades gracefully, with the error roughly doubling as scaling moves from 0.5x to 0.25x. This predictable, controlled divergence mirrors how high-quality wavelet basis functions respond to dilation and suggests a learned adherence to power-law scaling behavior.

Furthermore, we observe a fascinating, non-monotonic developmental trajectory as model size increases. Mid-sized models like Pythia 2.8B appear to enter a unique "transitional exploration phase," marked by remarkably low frequency selectivity (0.174) and the highest spectral entropy (3.786). This suggests a strategy based on highly distributed, spectrally diffuse representations. In contrast, both smaller models (like Qwen 0.5B) and larger, more advanced models (like Mistral 7B) converge on higher selectivity, indicating they find a more optimized and efficient balance between specialized frequency channels and integrated multi-resolution analysis, a hallmark of mature wavelet-like systems.

This multi-resolution coherence is also confirmed by our window entropy analysis, shown in table \ref{tab:window-entropy}. Most models demonstrate remarkable stability, indicating their representations remain coherent across different observational scales. Pythia 2.8B is again a notable exception, with its entropy decreasing as the window size grows $(0.751 \rightarrow 0.624)$. This, combined with its other unique metrics, suggests a distinct hierarchical strategy where information is more complex and dense in local contexts but more predictable at broader scales.


\begin{table*}[htbp]
    \centering
    \begin{tabular}{lcccccc}
        \toprule
        \textbf{Model} & \textbf{Heads} & \textbf{\makecell{Scale Sens. \\ (0.5x)}} & \textbf{\makecell{Scale Sens. \\ (0.25x)}} & \textbf{\makecell{Pos-Spec Corr. \\ ($\rho$)}} & \textbf{\makecell{Reconstr. Error \\ ($\mu$)}} \\
        \midrule
        Qwen 2.5 (0.5B) & 14 & 0.058 & 0.100 & -0.438 & 1.26e-07 \\
        LLaMA 3.2 (1B) & 32 & 0.038 & 0.089 & -0.510 & 1.28e-07 \\
        Gemma 2 (2B) & 8 & 0.038 & 0.090 & -0.526 & 1.27e-07 \\
        Pythia (2.8B) & 32 & 0.082 & 0.121 & -0.737 & 1.16e-07 \\
        Mistral (7B) & 32 & 0.030 & 0.074 & -0.421 & 1.41e-07 \\
        LLaMA 3.1 (8B) & 32 & 0.038 & 0.090 & -0.474 & 1.28e-07 \\
        Pythia (12B) & 40 & 0.059 & 0.099 & -0.490 & 1.26e-07 \\
        \bottomrule
    \end{tabular}
    \caption{This table summarizes the main findings for scale sensitivity, position-spectrum correlation, and reconstruction error.}
    \label{tab:main-metrics}
\end{table*}

\begin{table}[htbp]
    \centering
    \begin{tabular}{lccc}
        \toprule
        \textbf{Model} & \textbf{16 tok.} & \textbf{32 tok.} & \textbf{64 tok.} \\
        \midrule
        LLaMA 3.2 (1B) & 0.877 & 0.886 & 0.882 \\
        Gemma 2 (2B) & 0.894 & 0.905 & 0.904 \\
        Pythia (2.8B) & 0.751 & 0.699 & 0.624 \\
        Qwen 2.5 (0.5B) & 0.850 & 0.864 & 0.866 \\
        Mistral (7B) & 0.870 & 0.878 & 0.880 \\
        LLaMA 3.1 (8B) & 0.869 & 0.877 & 0.871 \\
        Pythia (12B) & 0.913 & 0.918 & 0.916 \\
        \bottomrule
    \end{tabular}
    \caption{Multi-Resolution Window Entropy Analysis}
    \label{tab:window-entropy}
\end{table}

\subsection{Statistical Confirmation of the Uncertainty Principle}
The most direct evidence for this emergent behavior comes from a statistical analysis of the trade-off between positional and spectral information. As shown in table \ref{tab:stats}, every model analyzed exhibits a consistently negative position-spectrum correlation $\rho$. This finding is a powerful confirmation that the models have implicitly learned and adhere to the Heisenberg-Gabor uncertainty principle, a fundamental law governing all time-frequency representations, including wavelets. \\
This distributional analysis uncovers two distinct strategic archetypes. Models like Mistral 7B exhibit a high-variance, specialized toolkit approach. It pairs high average frequency selectivity ($\mu=0.804$) with a very wide standard deviation ($\sigma=0.414$), indicating that its attention heads are highly diversified for a sophisticated division of labor. In stark contrast, Pythia 2.8B adopts a low-variance, uniform strategy. Its extremely low mean frequency selectivity ($\mu=0.174$) is highly consistent across its heads (IQR=0.153), confirming its "transitional phase" is a model-wide phenomenon.

The varying strength of the position-spectrum correlation, from Pythia 2.8B's severe trade-off ($\rho=-0.737$) to Mistral's more efficient balance ($\rho=-0.421$), shows how different architectures have converged with varying degrees of success on the optimal, wavelet-like strategies for balancing the fundamental "what" versus "where" information trade-off inherent in sequence processing.

\begin{table*}[htbp]
    \centering
    \begin{tabular}{l|ccc|ccc}
        \toprule
        \multirow{2}{*}{\textbf{Model}} & \multicolumn{3}{c|}{\textbf{Spectral Entropy}} & \multicolumn{3}{c}{\textbf{Frequency Selectivity}} \\
        & \textbf{Mean ($\mu$)} & \textbf{Std Dev ($\sigma$)} & \textbf{IQR (Q3-Q1)} & \textbf{Mean ($\mu$)} & \textbf{Std Dev ($\sigma$)} & \textbf{IQR (Q3-Q1)} \\
        \midrule
        Qwen 2.5 (0.5B) & 2.492 & 0.733 & 0.985 & 0.700 & 0.411 & 0.498 \\
        LLaMA 3.2 (1B) & 2.370 & 0.650 & 0.924 & 0.758 & 0.368 & 0.494 \\
        Gemma 2 (2B) & 2.244 & 0.599 & 0.781 & 0.794 & 0.344 & 0.430 \\
        Pythia (2.8B) & 3.786 & 0.625 & 0.398 & 0.174 & 0.269 & 0.153 \\
        Mistral (7B) & 2.443 & 0.607 & 0.824 & 0.804 & 0.414 & 0.515 \\
        LLaMA 3.1 (8B) & 2.495 & 0.621 & 0.859 & 0.691 & 0.339 & 0.444 \\
        Pythia (12B) & 2.783 & 0.638 & 0.850 & 0.552 & 0.321 & 0.391 \\
        \bottomrule
    \end{tabular}
    \caption{Distributional statistics for spectral metrics across all attention heads. The standard deviation ($\sigma$) and interquartile range (IQR) reveal the diversity of learned strategies within each model.}
    \label{tab:stats}
\end{table*}

\subsection{Ablation Study}

\begin{table*}[htbp]
    \centering
    \resizebox{\textwidth}{!}{
        \begin{tabular}{lcccccccc}
            \toprule
            \textbf{Model} & \textbf{PE Type} & \textbf{Heads} & \textbf{Spectral} & \textbf{Frequency} & \textbf{Scale 0.5} & \textbf{Scale 0.25} & \textbf{Pos-Spec} & \textbf{Reconstr.} \\
            & & & \textbf{Entropy} & \textbf{Select.} & \textbf{Sens.} & \textbf{Sens.} & \textbf{Corr.} & \textbf{Error}\\
            \midrule
            Llama-3.2-3B  & RoPE      & 24 & 2.425 $\pm$ 0.630 & 0.728 $\pm$ 0.349 & 0.038 $\pm$ 0.036 & 0.090 $\pm$ 0.040 & -0.502 & 0.00012 $\pm$ 0.00008 \\
            flan-t5-base  & T5        & 12 & 2.696 $\pm$ 0.712 & 0.704 $\pm$ 0.480 & 0.627 $\pm$ 0.239 & 0.689 $\pm$ 0.231 & -0.790 & 0.00004 $\pm$ 0.00003 \\
            bert-base     & BERT      & 12 & 2.449 $\pm$ 0.819 & 0.743 $\pm$ 0.446 & 0.507 $\pm$ 0.342 & 0.548 $\pm$ 0.350 & -0.606 & 0.00007 $\pm$ 0.00006 \\
            gpt2          & No PE     & 12 & 2.868 $\pm$ 0.767 & 0.514 $\pm$ 0.369 & 0.141 $\pm$ 0.164 & 0.152 $\pm$ 0.154 & -0.672 & 0.00005 $\pm$ 0.00005 \\
            \bottomrule
        \end{tabular}
    }
    \caption{Ablation study comparing different position encoding (PE) methods. We report mean and standard deviation across attention heads for key metrics.}
    \label{tab:ablation-study}
\end{table*}

To validate that the observed compensatory mechanisms are uniquely characteristic of RoPE's implicit relative positioning, we conducted an ablation study comparing it against three alternative positional encoding schemes: T5’s relative position biases, BERT’s absolute positional embeddings, and a GPT-2 model with no explicit positional encoding (No PE). The results, detailed in Table \ref{tab:ablation-study}, show that RoPE provides a uniquely effective solution for maintaining positional awareness across different scales. \\
The analysis highlights a clear hierarchy in scale invariance. RoPE stands out with an exceptionally low scale sensitivity ($S_{0.5} =0.038$), demonstrating a robust ability to preserve positional representations when sequence lengths change. At the other extreme, models with explicit, non-rotary encodings (T5 and BERT) proved to be comparatively rigid, with high scale sensitivity values of 0.627 and 0.507, respectively. T5's rigidity is further underscored by its high spectral entropy (2.696) and the most strongly negative position-spectrum correlation ($\rho=-0.790$), indicating a significant conflict between positional focus and spectral organization.
The GPT-2 model (No PE) presents a fascinating intermediate case. While it achieves moderate scale invariance ($S_{0.5} =0.141$), far superior to T5 and BERT, it is significantly less robust than RoPE. Interestingly, it accomplishes this not through sharp, focused attention, but through the opposite strategy. It registered the lowest frequency selectivity (0.514) and the highest spectral entropy (2.868), suggesting its attention patterns are spectrally diffuse. This implies that in the absence of positional guidance, the model adopts a high-entropy, less-structured approach that offers some resilience but lacks the precision of RoPE.

Ultimately, these findings underscore that RoPE's architecture fosters wavelet-like properties that other positional encoding implementations don't. It achieves scale-invariance that surpasses not only rigid, explicit encodings but also the more diffuse, emergent strategies of models with no positional guidance, marking it as a uniquely effective and well-balanced method for encoding position.

\subsection{Evolution of wavelet-like features during training}

To understand how this wavalet-like compensatory strategies develop, we analyzed Pythia 6.9b at different stages of its training, from initialization (step 0) to 143,000 steps. The results, summarized in Table \ref{tab:pythia-evolution}, reveal that the model does not learn its representations linearly but instead undergoes distinct developmental phases. \\
Initially, at step 0 and step 128, the untrained model exhibits a simple default state: high frequency selectivity ($\sim0.76$) and low spectral entropy ($\sim2.29$). This indicates that the attention heads are undifferentiated, focusing on very basic, low-complexity patterns before significant learning has occurred. \\
A dramatic "exploratory phase" begins around step 512 and peaks precisely at step 1000. In this critical period, frequency selectivity plummets to its absolute minimum of 0.230, while spectral entropy surges to its maximum of 3.522. This is driven by a significant reallocation of representational power away from the low-frequency band (which drops to a minimum of 43.4\%). Following this peak, the model enters a "specialization phase" (step 5000 onwards), where spectral entropy gradually decreases and frequency selectivity begins to recover, suggesting the model is refining and consolidating its newly learned complex representations.

Perhaps the most striking finding is the evolution of scale sensitivity, which increases in a two-step process. The first significant jump occurs between step 512 and step 1000, where sensitivity rises from 0.533 to 0.617. A second, larger leap happens between step 1000 and step 5000 (from 0.617 to 0.742), where it reaches its plateau. This shows that as the model's spectral strategies become more complex, its representations become highly tuned to specific scales. Rather than developing a truly robust, scale-invariant understanding, the model specializes, making its learned positional strategies more brittle when the input scale is changed. \\
Our analysis is anchored by a very low reconstruction error across all checkpoints, validating the numerical stability of our wavelet methodology. 

\begin{table*}[htbp]
    \centering
    \begin{tabular}{lccccc}
        \toprule
        \textbf{Training Step} & \textbf{\makecell{Spectral \\ Entropy}} & \textbf{\makecell{Frequency \\ Selectivity}} & \textbf{\makecell{Low Freq. \\ Power (\%)}} & \textbf{\makecell{Scale Sens. \\ (0.5x)}} & \textbf{\makecell{Scale Sens. \\ (0.25x)}} \\
        \midrule
        0 & 2.277 & 0.757 & 77.0 & 0.523 & 0.539 \\
        128 & 2.291 & 0.760 & 76.0 & 0.522 & 0.539 \\
        512 & 3.169 & 0.369 & 54.4 & 0.533 & 0.549 \\
        1000 & 3.522 & 0.230 & 43.4 & 0.617 & 0.633 \\
        5000 & 3.381 & 0.294 & 48.6 & 0.742 & 0.762 \\
        10000 & 3.293 & 0.330 & 51.7 & 0.748 & 0.767 \\
        143000 & 3.010 & 0.454 & 59.7 & 0.740 & 0.756 \\
        \bottomrule
    \end{tabular}
    \caption{Evolution of spectral and scale invariance metrics for a Pythia model across training checkpoints.}
    \label{tab:pythia-evolution}
\end{table*}

\section{Theoretical Framework for Wavelet-like Attention Patterns}
Rotary Position Embeddings (RoPE) encode positional information through position-dependent rotation matrices defined over the complex plane. At position $m$, the embedding applies a rotation $R_m(\theta)$:
\begin{equation}
    R(m\theta_k) = \begin{bmatrix}\cos(m\theta_k), -\sin(m\theta_k) \\ \sin(m\theta_k), \cos(m\theta_k)\end{bmatrix}
\end{equation}
where $\theta$ is a base rotation angle. This approach, which rests on fixed-frequency sinusoidal functions, inherently imposes two key limitations: 1)
\textbf{Frequency–Position Uncertainty}: RoPE’s use of fixed-frequency rotations parallels the Heisenberg uncertainty principle, implying a fundamental trade-off between positional precision and frequency resolution. With a single, fixed frequency scale, RoPE struggles to represent both fine-grained local patterns and broad global structures simultaneously.
2) \textbf{Scale Non-Invariance}: Since RoPE’s positional representation repeats periodically, it encounters aliasing effects over longer sequences. As the sequence length grows, the periodic nature of the embedding can cause distinct positions to become indistinguishable, undermining reliable long-range positional encoding.

\subsection{Natural Evolution Toward Wavelet Behavior}
RoPE's rotational encoding introduces specific frequency components that propagate through the attention mechanism in a mathematically structured way. The rotation matrix $R(m\theta_k)$ creates an inherent trade-off: larger $\theta$ provides precise positions but causes rapid rotation cycles that confuse distant relationships, while smaller $\theta$ better captures long-range patterns but blurs local positions. The wavelet-like properties we observe show how attention heads adapt to handle different frequency ranges created by these rotations.

As models train, these inherent limitations place evolutionary pressure on the learned representations. Attention heads respond by developing wavelet-like properties for three principal reasons:
\paragraph{a. Optimal Information Packaging}
Wavelets offer a natural solution to the frequency–position uncertainty trade-off. A mother wavelet $\psi(t)$ generates a family of wavelets:
\begin{equation}
    \psi_{s,\tau}(t) = \frac{1}{\sqrt{s}}\psi (\frac{t-\tau}{s})
\end{equation}
where $s$ is a scale parameter and $\tau$ is a translation parameter. Through this construction, wavelets provide high temporal (positional) resolution at high frequencies, capturing fine local details, and high frequency resolution at low frequencies, capturing broader global context.
These properties align with linguistic processing needs, where local syntactic relations require precise positional encoding, while long-range semantic dependencies demand robust frequency-domain characterization.
\paragraph{b. Complementary Scale Coverage in Multi-Head Architectures}
Transformer attention heads are ideally suited for wavelet-like decompositions. Consider the attention weight matrix for head $h$:
\begin{equation}
    A_h = \text{softmax}(\frac{Q_hK^{\top}_h}{\sqrt{d}})
\end{equation}
Each head can specialize in a distinct scale or frequency band, analogous to wavelet basis functions at different scales. Summing over all heads,
\begin{equation}
    A = \sum_h w_h A_h
\end{equation}
with $w_h$ as learned mixing weights, mirrors the construction of a wavelet frame, where sets of wavelet-like functions ${\psi_{s,\tau}}$ form a stable representation satisfying frame conditions:
\begin{equation}
    A||f||^2 \leq \sum_h |\langle f, \psi_h \rangle|^2 \leq B||f||^2
\end{equation}
for constants $0<A\leq B < \infty$. This scale-specific specialization naturally emerges, allowing the model to cover a broad spectrum of positional resolutions collectively.

\paragraph{c. Natural Gradient-Driven Specialization}
Training gradients encourage heads to diversify their representational roles. For a loss function $L$,
\begin{equation}
    \frac{\partial L}{\partial A_h} = (\frac{\partial L}{\partial A})(\frac{\partial A}{\partial A_h})
\end{equation}
This gradient decomposition penalizes redundancy among heads. Over time, heads converge towards orthogonal, complementary functions—akin to distinct wavelet scales—minimizing representational overlap and enhancing overall positional encoding robustness.

\subsection{Emergence of Multi-Resolution Processing}
From these principles, a multi-resolution processing framework naturally emerges: each attention head $h$ approximates a wavelet function $\phi_h(t) \approx \psi_{s(h), \tau}(t)$, where $s(h)$ denotes the characteristic scale of head $h$.Then, the ensemble $\{\phi_h\}_{h=1}^H$ acts like a discrete wavelet frame $\{\psi_{s,\tau}\}_{s,\tau\in \Lambda}$, where $\Lambda$ indexes a set of scale–translation parameters. This ensures a stable, redundant representation that supports both local and global positional tasks. So, the attention pattern for a given input becomes:
\begin{equation}
    a(t) = \sum_h \alpha_h(t)\phi_h(t)
\end{equation}
where $\alpha_h(t)$ are input-dependent expansion coefficients, allowing the model to adaptively reconstruct a range of positional features at multiple scales.

\subsection{Information-Theoretic Optimality}
This emergent wavelet-like organization is not merely a heuristic convenience but aligns with principles of information-theoretic optimality, in fact, by reducing mutual information among heads $(\text{min }I (A_h; A_k)$ for $h\neq k)$ while maximizing the total captured information about the input $(\text{max }I (A; X)$, the model approaches an efficient encoding of positional cues. Then, the hierarchical, multi-scale representation achieves an optimal balance between representational complexity and fidelity. Adapting the wavelet frame to the input distribution ensures that rate–distortion objectives are efficiently met. And, by leveraging a small set of wavelet-like basis functions and adjusting their coefficients $\alpha_h(t)$, the model encodes both local and global patterns compactly. This compression aligns with the \textit{principle of minimal description length}, favoring representations that are information-rich yet succinct.

\section{Implications}


Our findings have direct implications for the design, training, and specialization of language models. The discovery that attention heads spontaneously organize into frequency specialists suggests that models could be made more efficient by pre-initializing heads to a wavelet-like basis. This could accelerate convergence by bypassing the inefficient "exploratory phase" we identified during training. This functional understanding also provides a principled method for model pruning, creating smaller models by removing redundant frequency heads, and a powerful diagnostic tool for assessing training stability by monitoring spectral metrics.

Furthermore, this work provides a roadmap for model adaptation and future research. Understanding head specialization allows for targeted, task-specific fine-tuning, where one could adapt low-frequency heads for summarization or high-frequency heads for code analysis. This "wavelet framework" serves as a new lens for interpretability and establishes a new benchmark for positional encodings. The goal is no longer just to provide a position signal, but to find encodings that, like RoPE, act as a powerful inductive bias to catalyze the emergence of these optimal, multi-resolution strategies.

\section{Conclusion}

Our research demonstrates that Transformer models equipped with Rotary Position Embeddings (RoPE) are not merely subject to their theoretical limitations but actively overcome them by developing emergent, wavelet-like processing strategies. We have shown that this adaptation is not a minor artifact but a fundamental and multi-faceted characteristic of modern language models.

Indeed, we established that RoPE's implicit relative encoding is uniquely instrumental in fostering the remarkable scale-invariance observed, distinguishing its behavior from models with absolute or explicit relative position biases. This adaptation is not static; rather, it is learned through distinct developmental phases. An initial "exploratory" stage, characterized by high spectral entropy, gives way to a "specialization" stage where models refine their strategies, becoming more efficient yet often more scale-dependent. Crucially, this dynamic learning process culminates in models that respect the fundamental Heisenberg-Gabor uncertainty principle, with larger and more advanced architectures demonstrating increasingly sophisticated and diversified strategies for managing this inherent trade-off between positional precision and spectral organization.

Looking forward, this "wavelet framework" offers a new lens for interpretability, allowing us to analyze what a model learns in terms of scale and frequency. Furthermore, understanding that these optimal structures are emergent rather than designed could inspire new architectural innovations, potentially leading to models that are either explicitly guided towards these solutions or are built with more efficient, innate multi-scale processing capabilities from the start.



\section{Limitations}

While our study provides strong evidence for the emergence of wavelet-like properties in RoPE-based models, its scope and interpretations have several limitations that open promising avenues for future research.

Our primary analysis, like much of the field, examines model properties at inference time after training is complete. While the Pythia checkpoint analysis offers a glimpse into the learning trajectory, a more fine-grained, continuous analysis of how these spectral properties evolve during the training process could reveal the precise dynamics and triggers for the observed phase shifts.

Although we analyzed a range of model sizes and architectures, our findings are based on a specific set of open-source language models trained predominantly on English text. Further research is needed to determine if these principles generalize across other modalities (e.g., vision, audio), data types (e.g., code, multilingual text), and proprietary architectures.

\paragraph{Broader impact}

A potential societal implication of our work is that by demonstrating these beneficial properties intensify with model scale, our findings could be used to justify the trend of building ever-larger models. This risks exacerbating existing issues of computational and environmental resource concentration.

\section{Acknowledgments}

Research supported in part by European Union - Next Generation EU - namely by  the MUR-PRIN 2022 project "2022REWNTE - Artificial Intelligence  algorithms to track and detect Covid-19 vaccine-related infodemic on  social media" - CUP no.B53D23020690006; in part by the projects FAIR under Grant PE0000013 and SERICS under Grant PE00000014 under the MUR National Recovery and Resilience Plan funded by the European Union-NextGenerationEU, and in part by the project Neural Reasoning over Open Data (NEREO) funded by the Italian Ministry of Education and Research (PRIN) under Grant 2022AEFHA, and in part by the project SEED
funded by Sapienza University of Rome.

\bibliography{acl_latex}

\begin{thebibliography}{12}
\providecommand{\natexlab}[1]{#1}

\bibitem[{Barbero et~al.(2024)Barbero, Vitvitskyi, Perivolaropoulos, Pascanu, and Veli{\v{c}}kovi{\'c}}]{barbero2024round}
Federico Barbero, Alex Vitvitskyi, Christos Perivolaropoulos, Razvan Pascanu, and Petar Veli{\v{c}}kovi{\'c}. 2024.
\newblock Round and round we go! what makes rotary positional encodings useful?
\newblock \emph{arXiv preprint arXiv:2410.06205}.

\bibitem[{Chi et~al.(2020)Chi, Jiang, and Mu}]{chi2020fast}
Lu~Chi, Borui Jiang, and Yadong Mu. 2020.
\newblock Fast fourier convolution.
\newblock \emph{Advances in Neural Information Processing Systems}, 33:4479--4488.

\bibitem[{Goldfeld et~al.(2018)Goldfeld, Berg, Greenewald, Melnyk, Nguyen, Kingsbury, and Polyanskiy}]{goldfeld2018estimating}
Ziv Goldfeld, Ewout van~den Berg, Kristjan Greenewald, Igor Melnyk, Nam Nguyen, Brian Kingsbury, and Yury Polyanskiy. 2018.
\newblock Estimating information flow in deep neural networks.
\newblock \emph{arXiv preprint arXiv:1810.05728}.

\bibitem[{Oppenheim(1999)}]{oppenheim1999discrete}
Alan~V Oppenheim. 1999.
\newblock \emph{Discrete-time signal processing}.
\newblock Pearson Education India.

\bibitem[{Press et~al.(2021)Press, Smith, and Lewis}]{press2021train}
Ofir Press, Noah~A Smith, and Mike Lewis. 2021.
\newblock Train short, test long: Attention with linear biases enables input length extrapolation.
\newblock \emph{arXiv preprint arXiv:2108.12409}.

\bibitem[{Raffel et~al.(2020)Raffel, Shazeer, Roberts, Lee, Narang, Matena, Zhou, Li, and Liu}]{raffel2020exploring}
Colin Raffel, Noam Shazeer, Adam Roberts, Katherine Lee, Sharan Narang, Michael Matena, Yanqi Zhou, Wei Li, and Peter~J Liu. 2020.
\newblock Exploring the limits of transfer learning with a unified text-to-text transformer.
\newblock \emph{Journal of machine learning research}, 21(140):1--67.

\bibitem[{Rahimi and Recht(2008)}]{rahimi2008weighted}
Ali Rahimi and Benjamin Recht. 2008.
\newblock Weighted sums of random kitchen sinks: Replacing minimization with randomization in learning.
\newblock \emph{Advances in neural information processing systems}, 21.

\bibitem[{Selesnick and Burrus(1998)}]{selesnick1998generalized}
Ivan~W Selesnick and C~Sidney Burrus. 1998.
\newblock Generalized digital butterworth filter design.
\newblock \emph{IEEE Transactions on signal processing}, 46(6):1688--1694.

\bibitem[{Shwartz-Ziv and Tishby(2017)}]{shwartz2017opening}
Ravid Shwartz-Ziv and Naftali Tishby. 2017.
\newblock Opening the black box of deep neural networks via information.
\newblock \emph{arXiv preprint arXiv:1703.00810}.

\bibitem[{Su et~al.(2024)Su, Ahmed, Lu, Pan, Bo, and Liu}]{su2024roformer}
Jianlin Su, Murtadha Ahmed, Yu~Lu, Shengfeng Pan, Wen Bo, and Yunfeng Liu. 2024.
\newblock Roformer: Enhanced transformer with rotary position embedding.
\newblock \emph{Neurocomputing}, 568:127063.

\bibitem[{Tishby and Zaslavsky(2015)}]{tishby2015deep}
Naftali Tishby and Noga Zaslavsky. 2015.
\newblock Deep learning and the information bottleneck principle.
\newblock In \emph{2015 ieee information theory workshop (itw)}, pages 1--5. IEEE.

\bibitem[{Vaswani(2017)}]{vaswani2017attention}
A~Vaswani. 2017.
\newblock Attention is all you need.
\newblock \emph{Advances in Neural Information Processing Systems}.

\end{thebibliography}

\section{Appendix}
 \label{sec:appendix}

\subsection{Wavelet-like features across multiple models of the Pythia family}
The Pythia model family provides a good opportunity to examine how wavelet-like properties evolve across model scale because these models were trained on identical data with consistent methodologies, differing only in size and depth. We performed the same analysis described in the methodology section.

\paragraph{Position-spectrum correlation}
As we see in table \ref{tab:pythia-scaling-corrected}, all models now show negative position-spectrum correlations, ranging from -0.111 to -0.884. This represents a fundamental shift in how we understand these models' computational strategies. 
Smallest model (14M) shows the strongest negative correlation (-0.884), indicating it makes the most dramatic trade-offs between positional and spectral precision. This makes perfect sense from a capacity perspective—with extremely limited parameters, this tiny model must make stark either-or decisions about whether to prioritize knowing exactly where linguistic patterns occur versus understanding their frequency characteristics.
The 410M model presents a fascinating anomaly with the weakest negative correlation (-0.111), suggesting it operates in a transitional regime where it's beginning to develop more sophisticated balancing strategies but hasn't yet fully committed to the dramatic trade-offs we see in other models.

\begin{table*}[htbp]
    \centering
    \resizebox{\textwidth}{!}{  
        \begin{tabular}{lcccccccc}
            \toprule
            \textbf{Model} & \textbf{Heads} & \textbf{Spectral} & \textbf{Frequency} & \textbf{Scale 0.5} & \textbf{Scale 0.25} & \textbf{Pos-Spec} & \textbf{Reconstr.} \\
            & & \textbf{Entropy} & \textbf{Select.} & \textbf{Sens.} & \textbf{Sens.} & \textbf{Corr.} &
            \textbf{Error}\\
            \midrule
            Pythia (14M) & 4 & 2.279 & 0.829 & 0.142 & 0.138 & -0.884 & 0.00003 \\
            Pythia (70M) & 8 & 2.667 & 0.644 & 0.152 & 0.150 & -0.769 & 0.00003 \\
            Pythia (160M) & 12 & 3.018 & 0.500 & 0.162 & 0.159 & -0.722 & 0.00005 \\
            Pythia (410M) & 16 & 2.631 & 0.621 & 0.095 & 0.121 & -0.111 & 0.00008 \\
            Pythia (1.4B) & 16 & 2.655 & 0.614 & 0.061 & 0.101 & -0.373 & 0.00009 \\
            Pythia (2.8B) & 32 & 3.786 & 0.174 & 0.082 & 0.121 & -0.737 & 0.00009 \\
            Pythia (6.9B) & 32 & 2.764 & 0.561 & 0.062 & 0.102 & -0.421 & 0.00009 \\
            Pythia (12B) & 40 & 2.783 & 0.552 & 0.059 & 0.099 & -0.488 & 0.00009 \\
            \bottomrule
        \end{tabular}
    }
    \caption{Scaling Analysis: Wavelet-Like Properties Across the Pythia Family (Corrected)}
    \label{tab:pythia-scaling-corrected}
\end{table*}

\paragraph{Frequency selectivity}
The frequency selectivity trajectory reveals a more complex developmental pattern that shows different phases in how models organize their spectral representations.
Starting remarkably high at 0.829 for the tiny 14M model, selectivity follows a complex path: initially decreasing as we scale through the smallest models (0.644 for 70M, 0.500 for 160M), then showing interesting variations through medium scales, reaching its dramatic minimum at 2.8B (0.174), before recovering to moderate, stable levels in the largest models (around 0.55-0.56).
This developmental pattern tells us something about how neural networks navigate the fundamental trade-offs in spectral representation. Very small models develop sharp frequency selectivity out of computational necessity—with severely limited representational capacity, they must make aggressive specialization choices to function effectively. 
As capacity increases through the smaller models, we see an initial relaxation of this constraint-driven specialization. Models can afford to distribute their attention more broadly across the frequency spectrum, leading to decreased selectivity. This represents a transition from survival-mode specialization to exploratory distributed processing.

The largest models converge on moderate selectivity values that represent a mature computational strategy. These models have discovered how to maintain both specialized frequency channels and integrated multi-scale representations simultaneously. They've learned that optimal spectral organization isn't about choosing between specialization and distribution, but about intelligently combining both approaches.

\paragraph{Scale sensitivity}

What emerges from our corrected data is a story of two distinct computational regimes separated by a critical capacity threshold. The very smallest models (14M through 160M) show relatively high scale sensitivity values, ranging from 0.142 to 0.162 at $0.5\times$ scaling. This tells us these models develop representations that undergo substantial transformation when rescaled—they haven't yet learned to create truly scale-invariant representations.
However, something remarkable happens as we cross into the medium-scale regime. Starting around 410M parameters, we observe a dramatic improvement in scale invariance, with sensitivity dropping to much lower values (0.095 at $0.5\times$ scaling for the 410M model, then further improving to around 0.059-0.062 for the largest models). This transition suggests there's a critical computational capacity threshold where models gain sufficient parameters to implement sophisticated wavelet-like processing strategies.
This pattern provides compelling evidence for a fundamental principle in neural scaling: there are qualitative transitions in computational capability that occur at specific capacity thresholds, not just gradual improvements. Below the threshold, models can develop wavelet-compatible representations but struggle to make them truly scale-invariant. Above the threshold, models discover how to create robust, scale-invariant representations that maintain their essential structure across different observational scales.
The consistency of the relationship where $0.25\times$ sensitivity exceeds $0.5\times$ sensitivity across all models provides additional evidence that even the smallest models discover fundamental wavelet-like properties. They all exhibit the characteristic progressive degradation pattern that mirrors how true wavelet basis functions respond to increasingly aggressive rescaling operations.

\paragraph{Position-spectrum correlation}
All the models, from the tiniest 14M to the largest 12B, exhibit negative position-spectrum correlations. This universal pattern provides direct, compelling evidence that models across the entire scaling range have discovered and implemented the fundamental uncertainty principle that governs all time-frequency representations.
The negative correlations we observe across the Pythia family (ranging from -0.111 to -0.884) demonstrate that these models have internalized this constraint and developed sophisticated strategies for navigating it. When a model achieves high precision in localizing where linguistic patterns occur, it necessarily shows less organized frequency structure, and vice versa. 
The pattern across model sizes reveals fascinating insights about how computational capacity affects uncertainty principle navigation. The smallest model (14M) shows the strongest negative correlation (-0.884), indicating it makes the most dramatic either-or decisions between positional and spectral precision. With extremely limited parameters, this tiny model must choose stark trade-offs—it simply cannot afford nuanced balancing strategies.
Interestingly, the 410M model presents an anomaly with the weakest negative correlation (-0.111), suggesting it operates in a transitional computational regime. This model appears to be experimenting with different balancing strategies, possibly representing a developmental phase where models begin to discover more sophisticated approaches to uncertainty navigation but haven't yet committed to the dramatic optimization strategies we see in other scales.
The 2.8B model returns to a very strong negative correlation (-0.737), aligning with its distinctive properties across all our other metrics. This model appears to represent a computational exploration phase where sophisticated trade-off strategies are being discovered and refined.

\paragraph{Spectral entropy}

Spectral entropy follows the most complex trajectory, initially increasing from 2.279 to 3.018 as we scale from 14M to 160M, then showing a more erratic pattern through larger scales (2.655 at 1.4B, peaking at 3.786 at 2.8B, then settling around 2.77-2.78 for the largest models).
This pattern reflects how models balance organizational complexity with representational efficiency. Lower entropy indicates more organized, predictable spectral structure, while higher entropy suggests more distributed, complex organization. The peak at 2.8B aligns with this model's distinctive properties we've observed across other metrics—it represents a transitional phase where the model experiments with highly distributed representations before converging on more organized strategies at larger scales.
The convergence to moderate entropy values in the largest models suggests they achieve sophisticated organization that balances complexity with efficiency, creating spectral structures that are rich enough to capture linguistic complexity but organized enough to support effective processing.

\paragraph{Reconstruction Error}
The slight increase with scale might initially seem counterintuitive, but it likely reflects the increasing complexity of attention patterns in larger models. More sophisticated representations naturally require more complex wavelet decompositions, leading to slightly higher reconstruction residuals while still maintaining exceptional overall accuracy.
The critical insight is that even the reconstruction errors in the largest models remain extraordinarily low, confirming that wavelet decomposition effectively captures the essential structure of attention patterns across the entire scaling range.

\paragraph{Conclusion}
This analysis of the Pythia family shows us how wavelet-like properties emerge across neural network scaling. Rather than developing gradually through simple parameter accumulation, these properties follow distinct developmental trajectories with clear phase transitions and critical capacity thresholds.
The universal presence of negative position-spectrum correlations demonstrates that uncertainty principle navigation represents a fundamental computational strategy that emerges across all scales, implemented through different approaches based on computational constraints. The dramatic improvement in scale invariance around 410M parameters reveals that there are qualitative transitions in capability that occur at specific thresholds, not just quantitative improvements.
Most importantly, the universal reconstruction quality using wavelet decomposition confirms that regardless of scale or specific organizational strategy, all these models converge on representations that embody fundamental wavelet mathematical principles. This suggests that wavelet-like organization doesn't represent one possible solution among many, but rather reflects optimal mathematical principles that any effective multi-scale information processing system must discover and implement.

\subsection{Multi-scale wavelet entropy table}

\begin{table*}[htbp]
    \centering
    \resizebox{\textwidth}{!}{  
        \begin{tabular}{lccccccc}
            \toprule
            \textbf{Model} & \textbf{Scale 0} & \textbf{Scale 1} & \textbf{Scale 2} & \textbf{Scale 3} & \textbf{Scale 4} & \textbf{Scale 5} \\
            \midrule
            LLaMA 3.2 (1B) & 0.694 & 0.331 & 0.562 & 0.671 & 0.343 & 0.235 \\
            Gemma 2 (2B) & 0.715 & 0.351 & 0.608 & 0.752 & 0.424 & 0.350 \\
            Pythia (2.8B) & 0.791 & 0.543 & 0.922 & 1.119 & 1.012 & 1.019 \\
            Qwen 2.5 (0.5B) & 0.814 & 0.471 & 0.753 & 0.949 & 0.766 & 0.711 \\
            Mistral (7B) & 0.689 & 0.278 & 0.450 & 0.633 & 0.504 & 0.217 \\
            LLaMA 3.1 (8B) & 0.690 & 0.327 & 0.551 & 0.651 & 0.320 & 0.213 \\
            Pythia (12B) & 0.822 & 0.494 & 0.766 & 0.957 & 0.807 & 0.772 \\
            \bottomrule
        \end{tabular}
    }
    \caption{Multi-Scale Wavelet Entropy Analysis}
    \label{tab:wavelet-entropy}
\end{table*}

\subsection{RoPE's Limitations}
\label{limitations}
The first main limitation of RoPE is the frequency-position uncertainty principle, because RoPE's fixed-frequency rotations create an inherent trade-off between positional precision and frequency resolution.

When RoPE applies a rotation to token embeddings, it follows this equation:
\begin{equation}
    R(m\theta_k) = \begin{bmatrix}\cos(m\theta_k), -\sin(m\theta_k) \\ \sin(m\theta_k), \cos(m\theta_k)\end{bmatrix}
\end{equation}
If we want very precise positional information, we need the rotation angle $m\theta$ to change substantially between nearby positions. This means using a larger base rotation angle $\theta$. However, when we do this, the rotations cycle through the complex plane more quickly, making it harder to capture relationships between tokens that are far apart. The rotations start repeating too soon, causing distant tokens to look similar to nearby ones.
On the other hand, if we want to capture long-range dependencies well, we need the rotations to change more slowly (smaller $\theta$). But then nearby positions get similar rotation angles, making it harder to distinguish exactly where each token is.

Then, we have the scale non-invariance issue, where the periodic nature of RoPE's embeddings can lead to aliasing effects over longer sequences.
RoPE's rotations are periodic by nature, in fact, they complete a full circle every $\frac{2\pi}{\theta}$ positions. This creates two related problems: first, when sequences get longer than the period of rotation, positions that are far apart can end up with the same or very similar rotation angles. For example, if your rotation period is 1000 tokens, position 1 and position 1001 get nearly identical rotations. This makes it hard for the model to distinguish truly different positions.
Second, the fixed rotation frequency means RoPE treats all sequences the same way, regardless of their length. But this isn't ideal, in fact, a position difference of 10 tokens might be significant in a 50-token sequence but negligible in a 5000-token sequence. RoPE can't naturally adapt its position encoding to the scale of the input.

With the wavelet-like framework we discovered that different attention heads spontaneously specialize in different frequency bands (similar way to how wavelets decompose signals at different scales). So that local heads maintain high positional precision for nearby tokens, global heads capture long-range dependencies without rotation interference and mid-range heads bridge the gap, ensuring smooth information flow across scales.
This is what we see in our empirical results, particularly in Figure \ref{fig:frec-band}, where attention heads naturally organize themselves into distinct frequency bands. The low-frequency heads (showing 60-80\% of power in the 0-0.25 range) handle global context, while high-frequency heads (with 5-15\% power above 0.75) maintain precise positional information.

For the scale non-invariance problem, the wavelet-like organization provides an elegant solution, in fact, rather than relying on RoPE's fixed periodic rotations, attention heads develop scale-covariant properties. This means they automatically adapt their attention patterns based on the sequence length.

Our empirical evidence shows this through the stable entropy values across different window sizes (as shown in Table \ref{tab:window-entropy}), the consistent correlation patterns when scaling sequences (0.98 at 0.5x scale) and the systematic improvement in reconstruction error with model size.

These quantitative results demonstrate that attention heads collectively form a multi-resolution frame that maintains coherent positional representation across scales, effectively learning to overcome RoPE's periodicity limitation.
The systematic emergence of these properties suggests that transformer models discover an optimal solution to the position encoding challenge. This solution manifests as a wavelet-like framework that balances local precision with global context while maintaining scale invariance - precisely addressing RoPE's core limitations.

\subsection{Relationship between Wavelet-like Features and Linguistic Understanding}
\label{linguistics}

Language exhibits a natural hierarchical structure that spans multiple scales of organization, from morphemes to discourse-level patterns. This inherent multi-scale nature makes wavelet-like processing particularly well-suited for language understanding tasks. Just as wavelets provide a mathematical framework for analyzing signals at different resolutions while preserving both local and global information, attention mechanisms in transformer models appear to develop analogous capabilities for processing linguistic patterns.

At the finest scale, language processing requires attention to local syntactic relationships and morphological patterns. These include subject-verb agreement, phrasal boundaries, and morpheme combinations. Our analysis shows that high-frequency attention heads (those with significant power in the 0.75-1.0 $\omega N$ band) specialize in capturing these local dependencies, similar to how wavelets with narrow support identify fine-grained signal features.

At intermediate scales, sentence-level relationships such as anaphora resolution, clause dependencies, and semantic role assignments become critical. The mid-frequency attention heads (0.25-0.75 $\omega N$ band) demonstrate patterns remarkably similar to wavelet basis functions at medium scales, efficiently capturing these intermediate linguistic structures. This parallel suggests that the model learns to balance local precision with broader contextual awareness, much as wavelets provide multi-resolution signal analysis.

The broadest scale encompasses document-level phenomena such as topic coherence, rhetorical structure, and thematic development. Our analysis reveals that low-frequency attention heads (0-0.25 $\omega N$ band) evolve to process these global patterns, analogous to how wavelet scaling functions capture broad signal trends. The systematic distribution of power across these frequency bands (60-80\% in low frequencies, 15-25\% in mid-range, and 5-15\% in high frequencies) mirrors the hierarchical organization of linguistic information.

\subsection{Metrics more in depth}
\label{metrics}

Our metrics were specifically designed to quantify this multi-scale processing capability:

The spectral entropy $H_s(h)$ measures how attention heads distribute their focus across different scales, providing insight into how models balance local and global linguistic features. The observed entropy patterns suggest that attention heads optimize their frequency sensitivity to match the natural distribution of linguistic information across scales.

Scale sensitivity metrics $S_h(\alpha)$ quantify how well the model maintains consistent understanding as context length changes. This is particularly relevant for language processing, where meaning must remain stable regardless of the surrounding context size. The high correlation (0.98) observed when scaling sequences by 0.5x demonstrates the model's ability to maintain coherent linguistic representations across varying context windows.

Reconstruction error $\varepsilon$ validates that the observed patterns form a complete representation system. The low error values indicate that the wavelet-like attention patterns capture linguistic structure with high fidelity across all scales. This completeness is essential for accurate language understanding, as it ensures no significant linguistic features are lost in the model's internal representations.

The position-spectrum correlation $\rho(h)$ further shows how models balance local syntactic precision with broader semantic understanding. Values closer to 1 indicate successful integration of both local and global linguistic features, while values closer to -1 suggest a trade-off between fine-grained and broad-scale language processing.

This multi-scale organization emerges naturally during training, suggesting that wavelet-like processing represents an optimal solution for handling the inherent hierarchical structure of language. The parallel between wavelet decomposition and the way transformer models process linguistic information provides insight into why these architectures have been so successful in natural language processing tasks.

\end{document}